\documentclass[conference]{IEEEtran}
\usepackage{iftex}
\ifPDFTeX
  \usepackage{times}
\else
  \usepackage{fontspec}
\fi

\usepackage[numbers,compress]{natbib}
\usepackage{multicol}
\usepackage[bookmarks=true]{hyperref}

\usepackage{graphicx}
\usepackage{amsmath}
\usepackage{amssymb}
\usepackage{mathtools}
\usepackage{pgfpages}
\usepackage{pgf, tikz, adjustbox, pgfplots}
\usetikzlibrary{pgfplots.groupplots}
\pgfplotsset{compat=1.16} 
\usepackage{hyperref}
\usepackage[nameinlink, capitalize]{cleveref}
\usepackage{booktabs}
\usepackage[]{units}
\usepackage{epstopdf}
\usepackage{multirow}

\graphicspath{{figures/}}

\newcommand{\actangle}{\alpha}
\newcommand{\pendangle}{\varphi}
\newcommand{\kineticE}{T}
\newcommand{\potentialE}{U}
\newcommand{\lagrangian}{\mathcal{L}}

\newcommand{\gravity}{g}

\newcommand{\onemoment}{\eta}
\newcommand{\Pact}{\onemoment_\actangle} 
\newcommand{\Ppend}{\onemoment_\pendangle} 

\newcommand{\inertia}{J}
\newcommand{\Jact}{\inertia_\actangle}
\newcommand{\Jpend}{\inertia_\pendangle}
\newcommand{\Jcoup}{\inertia_{\actangle\pendangle}}

\newcommand{\torque}{\tau_\text{c}}
\newcommand{\torqueSP}{\tau_\text{c,SP}}

\newcommand{\pivottorque}{\boldsymbol{\tau}_\text{c}}

\newcommand{\jointtorque}{\tau_j}

\newcommand{\actpendangle}{\gamma}
\newcommand{\minactpendangle}{\gamma_{\min}}
\newcommand{\contiswitch}{s}

\newcommand{\torquesV}{\boldsymbol{\tau}}
\newcommand{\forcesV}{\boldsymbol{f}}
\newcommand{\dipoleM}{\mathbf{M}}
\newcommand{\jacM}{\mathbf{J}}
\newcommand{\actuationM}{\boldsymbol{\mathcal{A}}}

\newcommand{\magnetV}{\mathbf{\tilde{m}}}
\newcommand{\BField}{\boldsymbol{b}}
\newcommand{\BGrad}{\boldsymbol{g}}

\newcommand{\currentsV}{\boldsymbol{i}}
\newcommand{\currentsSP}{\currentsV_{\text{SP}}}
\newcommand{\positionV}{\boldsymbol{p}}

\newcommand{\zeroM}{\mathbf{0}}

\newcommand{\state}{x}
\newcommand{\xV}{\mathbf{\state}}
\newcommand{\tildexV}{\tilde{\xV}}
\newcommand{\superxV}{\boldsymbol{\state}}
\newcommand{\xtrajV}{\xV^*}
\newcommand{\xtauV}{\xV_\tau}

\newcommand{\outp}{y}

\newcommand{\superyV}{\boldsymbol{\outp}}

\newcommand{\uV}{\mathrm{u}}
\newcommand{\superuV}{\boldsymbol{u}}
\newcommand{\tildeuV}{\tilde{\uV}}
\newcommand{\utrajV}{\uV^*}

\newcommand{\liftoutM}{\mathbf{G}}
\newcommand{\liftinM}{\mathbf{F}}

\newcommand{\kalmanM}{\mathbf{K}}

\newcommand{\distV}{\boldsymbol{d}}
\newcommand{\distestV}{\boldsymbol{\widehat{d}}}

\newcommand{\scaleM}{\mathbf{S}}
\newcommand{\fdM}{\mathbf{D}}

\newcommand{\stateM}{\mathbf{A}}
\newcommand{\inputM}{\mathbf{B}}

\newcommand{\clstateM}{\bar{\stateM}}

\newcommand{\KtrajM}{\mathbf{K}_t}
\newcommand{\KeqM}{\mathbf{K}_\infty}

\renewcommand{\authorrefmark}[1]{\textsuperscript{#1}}

\hypersetup{
   pdfauthor={Viacheslav Sydora, Jasan Zughaibi, Denis von Arx, Quentin Boehler, Michael Muehlebach},
   pdftitle={Learning Dynamic Swing-Up of an Inverted Pendulum using Remote Magnetic Actuation},
   pdfsubject={Magnetic Actuation, Medical Robotics, Iterative Learning Control},
   pdfkeywords={electromagnetic navigation systems;iterative learning control;inverted pendulum;magnetic actuation;trajectory optimization;medical robotics}
}

\begin{document}

% paper title
\title{Learning Dynamic Swing-Up of an Inverted Pendulum using Remote Magnetic Actuation}

% \author{Author Names Omitted for Anonymous Review}
\author{\authorblockN{Viacheslav Sydora\authorrefmark{1}\authorrefmark{$\dagger$},
Jasan Zughaibi\authorrefmark{2}\authorrefmark{$\dagger$}\authorrefmark{*},
Denis von Arx\authorrefmark{2},
Quentin Boehler\authorrefmark{3},
Michael Muehlebach\authorrefmark{1}}
\\
\authorblockA{\authorrefmark{1}Learning and Dynamical Systems, Max Planck Institute for Intelligent Systems,
Tübingen, Germany}
\authorblockA{\authorrefmark{2}Multi-Scale Robotics Lab, Institute of Robotics and Intelligent Systems, D-MAVT, ETH Zürich,
Zürich, Switzerland}
\authorblockA{\authorrefmark{3}Medical Robotics Lab, Institute of Robotics and Intelligent Systems, D-MAVT, ETH Zürich,
Zürich, Switzerland}
\authorblockA{\authorrefmark{$\dagger$}These authors contributed equally.}
\authorblockA{\authorrefmark{*}Corresponding author. Email: \texttt{zjasan@ethz.ch}}}

\maketitle

\begin{abstract}
Electromagnetic Navigation Systems (eMNS) have gained considerable attention for minimally invasive surgery and targeted drug delivery. While most of the literature relies on quasi-static control of these systems, recent work has demonstrated the benefits of dynamic approaches. However, trajectory tracking far from equilibrium states remains largely unaddressed. We close this gap by demonstrating the first swing-up of a magnetically actuated inverted pendulum using the clinically-ready \textit{Navion} eMNS. Although the inverted pendulum is not clinically relevant in itself, the proposed method utilizes torques and forces as control objectives, making it applicable to other magnetically actuated devices such as catheters and guidewires. Our approach combines trajectory optimization that accounts for internal eMNS dynamics with time-varying Linear Quadratic Regulator (LQR) state feedback and Iterative Learning Control (ILC), which leverages previous trial data and the system’s dynamic model to progressively refine the feedforward command. While LQR alone fails due to the complex phenomena of magnetic actuation, ILC enables successful swing-up within six iterations. Furthermore, post-experimental analysis reveals that the learned ILC correction closely matches the torque discrepancy predicted by high-fidelity magnetic field model calibration, suggesting learning and adaptation as a promising tool to deal with uncertainties in electromagnetic actuation arising, e.g., from patient-specific physiological motion patterns and field model calibration inaccuracies.
\end{abstract}

\begin{IEEEkeywords}
electromagnetic navigation systems, iterative learning control, inverted pendulum, magnetic actuation, trajectory optimization, medical robotics
\end{IEEEkeywords}

\IEEEpeerreviewmaketitle

\section*{Supplementary Material}
A video featuring the swing-up demonstration is available at \url{https://youtu.be/1T4FVMYlteo}. The code and data accompanying this paper can be found at \url{https://github.com/slavasg-lab/emns_invpend_swingup}.

\section{Introduction}

Magnetic navigation systems enable wireless control of magnetic devices ranging from nanoscale to centimeter-sized objects through the modulation of magnetic fields, making them particularly attractive for medical robotics and applications spanning catheter steering, capsule endoscopy, and targeted drug delivery \cite{landers25, kim22, mesot25, heemeyer25, hwang20}.

Field generation in medical robotics takes two principal forms. Permanent-magnet systems rely on the mechanical motion of strong magnets and produce high field strengths and gradients, but are limited in responsiveness by mechanical inertia of the magnets. Electromagnetic Navigation Systems (eMNS), by contrast, generate fields through current-driven coils, enabling rapid modulation \cite{yang20} with bandwidths sufficient even for levitation of macroscopic objects \cite{singh26}.

While most of the literature neglects this advantage by relying on quasi-static modeling \cite{landers25, kim22, mesot25, heemeyer25, hwang20}, recent work demonstrates that dynamic approaches enable superior trajectory tracking and disturbance rejection, benefiting procedures such as intracardiac ablation \cite{zughaibi25a}. Moreover, real-time state feedback reduces required currents and substantially expands the operational workspace, enabling manipulation at distances relevant for human-scale clinical procedures \cite{zughaibi25b}.
% and enables levitation of macroscopic objects, opening new possibilities for ingestible capsule diagnostics \cite{singh26}.

\begin{figure}[!t]
    \centering
    \includegraphics[width=\columnwidth]{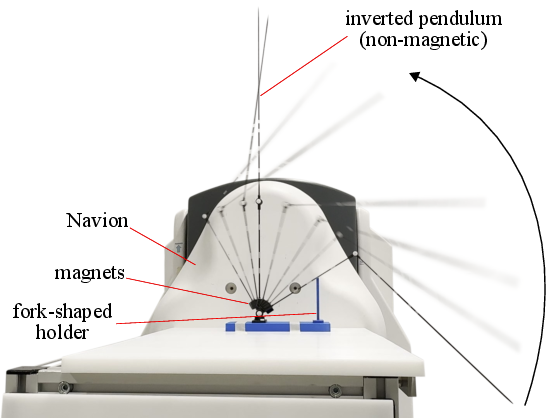}
    \caption{Stroboscopic visualization of the magnetically actuated inverted pendulum executing a swing-up maneuver using the clinically-ready Navion electromagnetic navigation system. The system comprises a magnetically-actuated lower arm (actuator) and a passive upper arm (pendulum) connected using a revolute joint. The sequence captures the transition from the downward position through the dynamic swing-up to the upright configuration, demonstrating successful trajectory tracking achieved through iterative learning control.}
    \label{fig:stroboframes}
\end{figure}

We build on these advances to further push the dynamic limits of eMNS. Similarly to previous work \cite{zughaibi25a, zughaibi25b}, we utilize an inverted pendulum as a benchmark, which is a well-established testbed for novel control architectures \cite{schoellig09, jung04, vu21} with numerous practical implementations \cite{vu21, hehn11, hofer18, gillespie16, gajamohan13, nagarajan14}. Although not clinically relevant in itself, the inverted pendulum is well-suited to explore advanced dynamic magnetic control strategies \cite{zughaibi25a, zughaibi25b}. The stabilization of the inverted pendulum is directly analogous to balancing a pole on the tip of a finger, which requires continuous control effort to counteract the inherent instability of the upright position. We address an even more challenging problem: the swing-up maneuver, which is illustrated in \cref{fig:stroboframes}. This maneuver involves driving a passive, non-magnetic pendulum using a magnetically actuated arm from the downward position to the upright equilibrium in a single continuous motion, concluding with a successful catch and stabilization in the upright position. This highly dynamic maneuver --- swing-up in \unit[0.6]{s} on the clinically-ready \textit{Navion} eMNS --- requires precise trajectory tracking and showcases a largely unexplored capability of eMNS. To highlight that our methodology is platform-agnostic, we additionally evaluated the swing-up on the OctoMag \cite{kummer10}, a research-grade, eight-coil system originally developed for ophthalmic surgery.

We achieve dynamic maneuvers through an optimization-based trajectory planner that explicitly accounts for eMNS actuation dynamics, paired with a finite-horizon discrete-time Linear Quadratic Regulator (LQR) for trajectory tracking. However, systematic model mismatches arising from imperfect magnetic field model calibration and inverted pendulum model uncertainties render a purely model-based approach insufficient. We therefore employ Iterative Learning Control (ILC), which combines data from previous trials with a model-based approach to learn a correction signal for the next iteration, much like a gymnast refining a vault based on the errors of past attempts. Notably, post-experimental analysis reveals that the learned ILC correction closely matches the torque discrepancy predicted by a subsequent high-fidelity magnetic field model calibration \cite{petruska17, bernardes26}. This suggests learning and adaptation as a promising tool to deal with uncertainties in electromagnetic actuation, which is particularly relevant in clinical settings, where physiological motion and patient-specific anatomy are difficult to model explicitly \cite{cagneau07,wang24,bebek07}.

The remainder of this paper is organized as follows. \cref{sec:related_work} reviews the related literature, while \cref{sec:platform} describes the experimental platform and the eMNS employed. \cref{sec:modeling} then derives the system dynamics and magnetic actuation models, after which \cref{sec:control} details the trajectory planning, feedback control, and ILC algorithms. Finally, \cref{sec:results} presents and discusses the experimental results, and \cref{sec:conclusion} concludes the paper.

\section{Related work}
\label{sec:related_work}
While most of the literature relies on quasi-static modeling of eMNS (see \cite{zughaibi25a} and references therein), recent works have demonstrated the benefits of dynamic approaches. In \cite{zughaibi25a}, eMNS dynamic capabilities were demonstrated by stabilizing a non-magnetic inverted pendulum, arguing that the available actuation bandwidth is sufficient to reject disturbances relevant to cardiovascular interventions. Building on this, the feedback control algorithms were extended to multi-agent control and energy-efficient field allocation, demonstrating that real-time state information can substantially expand the operational workspace \cite{zughaibi25b}. Most recently, the authors of \cite{singh26} presented levitation of a macroscopic object inside an eMNS, proposing applications in diagnostic procedures using sensorized ingestible capsules. Despite these advances, highly dynamic trajectory tracking far from equilibrium on eMNS remains unaddressed.

The inverted pendulum swing-up is a canonical benchmark for testing control architectures, as it requires driving a system through a non-equilibrium transient to an unstable equilibrium \cite{boubaker12}. Classical energy-based methods \cite{astrom96} addressed this through carefully designed switching laws, while data-driven approaches combined learned policies with nonlinear model predictive control to enforce constraints \cite{turrisi20,sferrazza20}. A persistent challenge across these methods is compensating for model mismatch. ILC addressed this effectively in the swing-up setting, demonstrating convergence in few trials by iteratively refining the feedforward command \cite{schoellig09, zughaibi21}.

ILC is a powerful tool for systems and tasks that are inherently repetitive \cite{bristow06}. In magnetic systems, ILC was applied to rate-dependent hysteresis compensation in shape memory alloy actuators \cite{yu22} and trajectory tracking in spherical permanent-magnet actuators \cite{zhang16}, demonstrating its ability to compensate for magnetic field modeling errors -- a challenge directly relevant to eMNS control. In clinical settings, physiological motions such as heartbeat and respiration induce quasi-periodic disturbances that are difficult to model explicitly but consistent across cycles. This property was leveraged in \cite{cagneau07} to compensate for periodic respiratory motion in force-controlled minimally invasive robotic surgery, as the complexity of tool-organ contact dynamics renders purely model-based approaches infeasible. In \cite{wang24}, ILC was applied to position control of a cable-driven soft robotic arm tracking irregular heartbeat on a simulated heart platform, demonstrating effective compensation for periodic lag caused by material compliance and system delays.

\section{Experimental Platform}
\label{sec:platform}
In this work, a custom-built inverted pendulum is employed, consisting of a magnetically actuated lower arm, which we refer to as the \textit{actuator}, and a passive, non-magnetic upper arm, which we refer to as the \textit{pendulum}. Both arms are made of non-magnetic, commercially available carbon-fiber reinforced thermoplastics. Three axially magnetized permanent magnets (supermagnete\textregistered{} R-20-04-05-N, NdFeB N45, $\unit[\varnothing~20~(4.2) \times 5]{mm}$) are mounted proximal to the actuator's pivot point, as shown in \cref{fig:schematic}. The actuator is mounted on a base plate using a custom non-magnetic revolute joint made from 3D-printed plastic, permitting rotation in one angular direction parametrized by $\actangle$. We use a similar joint to connect the actuator to the pendulum, providing an additional angular degree of freedom $\pendangle$. Relative motion between the two arms is constrained by a mechanical hard stop, which limits the maximum relative angular deflection to approximately $104^\circ$. Prior to swing-up, the inverted pendulum rests in the initial position against a fork-shaped holder, as illustrated in \cref{fig:stroboframes}, with the mechanical hard stop engaged at movement onset.

To demonstrate the platform-agnostic nature of our approach, we employ two distinct eMNS platforms to magnetically control the inverted pendulum system, showcasing its applicability across varied electromagnetic setups:
\begin{itemize}
	\item \textit{Navion}: a clinically-ready eMNS featuring three lateral coils in a triangular configuration \cite{gervasoni24}, as shown in \cref{fig:stroboframes}. Currents $\currentsV \in \mathbb{R}^3$ are sent to the driver at $f_\text{s}=\unit[100]{Hz}$ with a conservative safety limit of \unit[13]{A} per coil, well below the hardware's \unit[45]{A} capability, which would enable positioning the pendulum at greater distances from the coils. The system exhibits an actuation bandwidth $f_\text{b}$ of approximately \unit[25]{Hz}, defined as the corner frequency of the closed-loop electrical dynamics, beyond which the magnetic field response to control commands begins to degrade significantly \cite{zughaibi25b}.
	\item \textit{OctoMag}: a research-oriented eMNS comprising eight electromagnets \cite{kummer10}. Control currents $\currentsV \in \mathbb{R}^8$ are commanded at a frequency of $f_\text{s}=\unit[200]{Hz}$, with a safety limit imposed at \unit[4]{A} per coil. The system's actuation bandwidth was identified as $f_\text{b}=\unit[26.4]{Hz}$ in \cite{zughaibi25a}.
\end{itemize}

To enable real-time feedback, reflective markers are attached to both arms and tracked by an infrared motion capture system, with the sampling rate matched to the control frequency of the eMNS in use.

\section{Modeling}
\label{sec:modeling}
This section derives the analytical models governing the system dynamics and actuation. We first formulate the mechanical equations of motion for the inverted pendulum, then establish the generation of the magnetic wrench by the eMNS coils. Finally, we formulate the allocation problem to map the desired control torque to the desired coil currents.

\subsection{Inverted Pendulum Dynamics}

Following \cite{zughaibi25a}, the equations of motion are derived using the Lagrangian formalism with generalized coordinates ($\actangle$, $\pendangle$). The mechanical and electromagnetic subsystems are treated as decoupled \cite{zughaibi25b} by modeling the magnetic interaction as an external torque $\torque$, rather than including the magnetic potential energy in the Lagrangian. Thus, this paragraph focuses purely on mechanical modeling, while the electromagnetic actuation is treated in the subsequent paragraphs.

\begin{figure}
    \centering
    \includegraphics[width=\columnwidth]{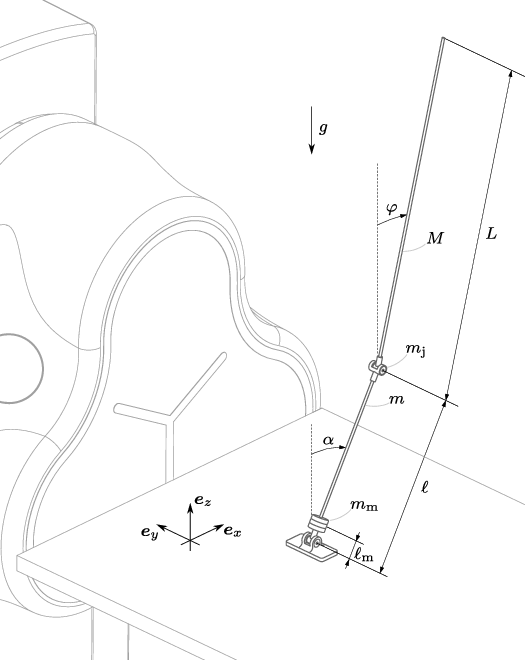}
    \caption{Schematic of the inverted pendulum platform showing the system parametrization and rotations relative to the motion capture inertial frame. Both the actuator and pendulum have a single rotational degree of freedom about the $\boldsymbol{e}_y$ axis, denoted by $\actangle$ and $\pendangle$, respectively. Mass parameters: actuator rod $m$, pendulum rod $M$, revolute joint $m_\text{j}$, magnets $m_\text{m}$. Length parameters: actuator $\ell$, pendulum $L$, and distance from the center of the magnetic volume to the actuator pivot point $\ell_\text{m}$.}
    \label{fig:schematic}
\end{figure}

The Lagrangian of the mechanical system is defined as $\lagrangian \coloneqq \kineticE - \potentialE$, where the kinetic energy $T$ and potential energy $U$ are given by:
\begin{align}
    \begin{aligned}
        T &= \frac{1}{2} \Jact \dot{\actangle}^2 + \frac{1}{2} \Jpend \dot{\pendangle}^2 + \Jcoup \dot{\actangle} \dot{\pendangle} \cos(\actangle-\pendangle) \\
        U &= \Pact \gravity \cos{\actangle} + \Ppend \gravity \cos{\pendangle}.
    \end{aligned}
\label{eq:lagrangian_energies}
\end{align}
Here, $\Pact$ and $\Ppend$ (in $\text{kg m}$) represent the first and  $\Jact$, $\Jpend$, and $\Jcoup$ (in $\text{kg m}^2$) the second moments of mass, respectively:
\begin{align}
\Jact &= m_\text{m} \ell_\text{m}^2 + \frac{1}{4} m \ell^2 + m_\text{j} \ell^2 + M L^2, \\
\Jpend &= \frac{1}{3} M L^2, & \Jcoup &= \frac{1}{2}M \ell L, \\
\Pact &= M \ell + m_\text{j} \ell + m_\text{m} \ell_\text{m} + \frac{1}{2} m \ell, & \Ppend &= \frac{1}{2} M L,
\end{align}
with mass and length parameters defined in \cref{fig:schematic}.

The differential equations governing the generalized coordinates $\actangle$ and $\pendangle$ are obtained from the Euler-Lagrange equations:
\begin{align}
\begin{aligned}
    \frac{\mathrm{d}}{\mathrm{d}t}\left(\frac{\partial\lagrangian}
    {\partial\dot{\actangle}}\right)
    - \frac{\partial\lagrangian}{\partial\actangle}
    &= \torque + \jointtorque \\
    \frac{\mathrm{d}}{\mathrm{d}t}\left(\frac{\partial\lagrangian}
    {\partial\dot{\pendangle}}\right)
    - \frac{\partial\lagrangian}{\partial\pendangle}
    &= - \jointtorque ,
\end{aligned}
\end{align}
where $\torque$ is the active torque acting on the actuator and $\jointtorque$ is the non-conservative torque that models the mechanical hard stop imposing a unilateral constraint $\actpendangle \geq \minactpendangle$ on the relative angle $\actpendangle = |\pi - (\actangle - \pendangle)|$ between the actuator and the pendulum.

To enable gradient-based trajectory optimization, the unilateral hard stop is approximated as a smooth, dissipative soft constraint. The torque is active only when the limit is violated ($\actpendangle < \minactpendangle$) with approaching velocity $\dot{\actpendangle} < 0$:
\begin{align}
    \jointtorque = d\dot{\actpendangle}[\contiswitch(\minactpendangle - \actpendangle) \contiswitch(\dot{\actpendangle})],
\end{align}
where $d > 0$ is the damping coefficient and $\contiswitch: \mathbb{R} \to [0,1]$ is a continuous switching function defined as
\begin{align}
    \contiswitch(a) \coloneqq \frac{\arctan(va)}{\pi} + \frac{1}{2},
\end{align}
with $v > 0$ being a steepness parameter that controls the sharpness of the transition. 

The resulting nonlinear dynamics are given by:
\begin{equation}
    \begin{split}
    &\begin{bmatrix}
        \Jact & \Jcoup \cos(\actangle - \pendangle) \\
        \Jcoup \cos(\actangle - \pendangle) & \Jpend
    \end{bmatrix}
    \begin{bmatrix}
        \ddot{\actangle} \\
        \ddot{\pendangle}
    \end{bmatrix}
    = \\
    & \begin{bmatrix}
        -\dot{\pendangle}^2 \Jcoup \sin(\actangle - \pendangle) + \Pact \gravity \sin(\actangle) + \jointtorque + \torque \\
        \phantom{-}\dot{\actangle}^2 \Jcoup \sin(\actangle - \pendangle) + \Ppend \gravity \sin(\pendangle) - \jointtorque
    \end{bmatrix}.
    \end{split}
    \label{eq:actpend_eom_compact}
\end{equation}

\subsection{Relation between Magnetic Wrench and Magnetic Field}

% This subsection establishes the magnetic actuation model relating eMNS coil currents to the mechanical wrench acting on the magnets. We first derive the force and torque experienced by a dipole in an external magnetic field, followed by the model for field generation by the eMNS coils.

A magnetic dipole with moment $\magnetV$ (in $\unit[]{Am^2}$), subject to an external magnetic field $\BField$, experiences a wrench comprising a torque $\torquesV = \magnetV \times \BField$ and a force $\forcesV = (\magnetV \cdot \nabla)\BField$, which can be written as a linear map \cite{petruska15}: 
\begin{align}
\begin{bmatrix}
    \torquesV \\
    \forcesV
\end{bmatrix}
=
\underbrace{
\begin{bmatrix}
    \dipoleM_{\BField}(\actangle) & \zeroM \\
    \zeroM & \dipoleM_{\BGrad}(\actangle)
\end{bmatrix}
}_{\dipoleM(\actangle)}
\begin{bmatrix}
    \BField \\
    \BGrad
\end{bmatrix}.
\label{eq:magnetic_wrench}
\end{align}
Here, the nine degrees of freedom of the magnetic field gradient $\nabla \BField$ are reduced to five independent components by applying Maxwell's equations under the assumption of a quasi-static field ($\nabla \cdot \BField = 0$) with no free currents ($\nabla \times \BField = \mathbf{0}$). These independent components are organized into the gradient vector $
    \BGrad \coloneqq 
    \begin{bmatrix}
        \frac{\partial b_x}{\partial x} &
        \frac{\partial b_x}{\partial y} &
        \frac{\partial b_x}{\partial z} &
        \frac{\partial b_y}{\partial y} &
        \frac{\partial b_y}{\partial z}
    \end{bmatrix}^\intercal
$ by convention. The skew-symmetric operator $\dipoleM_{\BField}$ and the gradient-to-force mapping matrix $\dipoleM_{\BGrad}$ are defined as:
\begin{align}
    \dipoleM_{\BField} &= \text{skew}(\magnetV) = \begin{bmatrix}
    0 & - \tilde{m}_z & \tilde{m}_y \\
    \tilde{m}_z & 0 & - \tilde{m}_x \\
    - \tilde{m}_y &  \tilde{m}_x & 0
\end{bmatrix}, \\
    \dipoleM_{\BGrad}  &= \begin{bmatrix}
    \tilde{m}_x & \tilde{m}_y & \tilde{m}_z & 0 & 0 \\
    0 & \tilde{m}_x & 0 & \tilde{m}_y & \tilde{m}_z \\
    -\tilde{m}_z & 0 & \tilde{m}_x & -\tilde{m}_z & \tilde{m}_y
\end{bmatrix}.
\end{align}

\subsection{Relation between Magnetic Field and Coil Currents}

Having established the relationship between torques/forces and field/gradients, we now model the fields and gradients caused by the electrical currents. The magnetic field and gradient vectors are commonly assumed to be linear functions of the coil currents $\currentsV \in \mathbb{R}^n, n \in \{3, 8\}$:
\begin{align}
    \begin{bmatrix}
    \BField \\
    \BGrad
\end{bmatrix}
= \actuationM(\positionV)\currentsV,
\label{eq:field_generation}
\end{align}
where the actuation matrix $\actuationM(\positionV) \in \mathbb{R}^{8 \times n}$ encodes the spatial field distribution and is evaluated in real-time at the dipole position $\positionV$ using a calibrated nonlinear multipole expansion \cite{petruska17}, with calibration procedures detailed in \cite{bernardes26}. This linear mapping relies on the principle of superposition, valid under the assumption that the electromagnetic cores operate far from magnetic saturation, which holds throughout our experiments.

\subsection{Relation between Magnetic Wrench and Control Torque}

The magnetic wrench generates mechanical torque through two mechanisms: the magnetic torque $\torquesV$ acting directly on the dipole, and the moment created by the magnetic force $\forcesV$ applied at lever arm distance $\ell_m$ from the pivot. We impose zero torque orthogonal to the swing up direction and leave the $z$-axis unconstrained to avoid an ill-conditioned system, as a magnetic dipole cannot generate torque about its own alignment direction at the upright equilibrium. This defines the target torque vector $\pivottorque \in \mathbb{R}^2$ as:
\begin{align}
    \pivottorque = \begin{bmatrix} 0 \\ \torque \end{bmatrix} = \jacM(\actangle) \begin{bmatrix} \torquesV \\ \forcesV \end{bmatrix},
\label{eq:torqueforce_projection}
 \end{align}
where the Jacobian $\jacM \in \mathbb{R}^{2 \times 6}$ projects the magnetic wrench onto the pivot's $x$- and $y$-axes:
\begin{align}
\jacM= \begin{bmatrix}
    1 & 0 & 0 & 0 & -\ell_m \cos \actangle & 0 \\
    0 & 1 & 0 & \ell_m \cos \actangle  & 0 & -\ell_m \sin \actangle
\end{bmatrix}.
\end{align}

\subsection{Allocation Problem}

By composing \eqref{eq:magnetic_wrench}, \eqref{eq:field_generation}, and \eqref{eq:torqueforce_projection}, we obtain the complete mapping from coil currents to the desired pivot torque:
\begin{align}
    \pivottorque = \jacM (\actangle) \dipoleM (\actangle) \actuationM(\positionV) \currentsV.
\end{align}
Consequently, the desired coil currents $\currentsSP$ are computed using the Moore-Penrose pseudoinverse \cite{zughaibi25b}:
\begin{align}
   \currentsSP = \left[ \jacM (\actangle) \dipoleM (\actangle) \actuationM(\positionV) \right]^\dagger \begin{bmatrix} 0 \\ \torqueSP \end{bmatrix},
\end{align}
where the torque setpoint $\torqueSP$ comprises the commands from the ILC and LQR controller detailed in \cref{sec:control}. Since the framework operates on torque and force objectives, it is hardware-agnostic and transfers directly to other eMNS platforms as well as other magnetically actuated devices such as catheters and guidewires. Note that this allocation requires real-time measurement of $\actangle$ and $\positionV$. Furthermore, in this work, we distinguish between the torque setpoint $\torqueSP$ and the actual torque $\torque$ acting on the system, as the internal eMNS dynamics prevent instantaneous tracking of the requested torque command.

\section{Control}
\label{sec:control}

\begin{figure*}[!t]
    \centering
    \includegraphics[width=\textwidth]{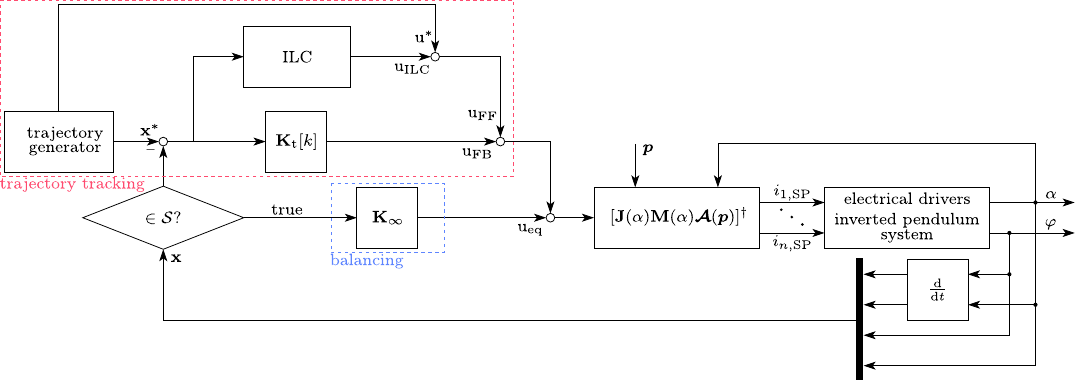}
    \caption{Block diagram of the control architecture. The system operates in two distinct modes: a trajectory tracking phase (red dashed box) for the swing-up maneuver, which combines time-varying LQR feedback $\KtrajM[k]$ with ILC, and a balancing phase (blue dashed box) that engages a linear regulator $\KeqM$ once the state enters the empirically defined region of attraction $\mathcal{S}$. The resulting torque command is mapped to optimal coil currents using the real-time allocation matrix $[\jacM (\actangle) \dipoleM (\actangle) \actuationM(\positionV)]^\dagger$.}
    \label{fig:control_loop}
\end{figure*}

The swing-up maneuver is executed as a sequence of two distinct control phases: trajectory tracking and balancing. Initially, the controller tracks a precomputed trajectory to guide the system from the downward configuration toward the upright position. Upon entering a predefined region of attraction around the upright equilibrium, the system switches to a balancing controller to regulate the unstable upright state, as illustrated in \cref{fig:control_loop}.

In the following section, we begin by providing the definition of the system, followed by the trajectory planning, feedback tracking, and ILC employed during the trajectory tracking phase, and conclude with the balancing controller.
% Initially, the controller tracks a precomputed trajectory to guide the system from the initial configuration toward the upright position. Upon entering a predefined region of attraction around the upright equilibrium, the system switches to a balancing controller to regulate the unstable upright state, as illustrated in the block diagram in \cref{fig:control_loop}.

\subsection{System Definition}

To synthesize the controllers described in the following paragraphs, we derive a linearized model of the system. While the full nonlinear dynamics described in \eqref{eq:actpend_eom_compact} are utilized for trajectory planning, controller synthesis relies on a simplified model where the joint constraint is omitted ($\jointtorque = 0$) and treated as an unmodeled disturbance.

We characterize this simplified system using the state vector $\xV \coloneqq [\actangle, \pendangle, \dot{\actangle}, \dot{\pendangle}]^\intercal$ and the scalar control input $\uV \coloneqq \torqueSP$, corresponding to the torque setpoint along the $y$-axis. The simplified dynamics are then linearized about the reference state and input trajectories $\xtrajV$ and $\utrajV$, respectively. The resulting time-varying system is discretized using a zero-order hold with a sampling time $T_\text{s} = 1/f_\text{s}$, yielding the discrete-time error dynamics:
\begin{equation}
	\tildexV[k+1] = \stateM[k] \tildexV[k] + \inputM[k] \tildeuV[k],
    \label{eq:discretized}
\end{equation}
where $\tildexV \coloneqq \xV - \xtrajV$ and $\tildeuV \coloneqq \uV - \utrajV$. The matrices $\stateM[k] \in \mathbb{R}^{n_x \times n_x}$ and $\inputM[k] \in \mathbb{R}^{n_x \times n_u}$ represent the discrete-time time-varying state and input system matrices, respectively.

\subsection{Trajectory Generation}
To generate feasible trajectories that respect the bandwidth limitations of the eMNS, we explicitly model the actuation dynamics arising from the low-level current control loop. Since the coil currents cannot be tracked instantaneously, the resulting magnetic torque $\torque$ lags behind the commanded setpoint $\uV$. We model this behavior using a first-order dynamical system:
\begin{align} \dot{\tau}_\text{c} = \omega_\text{b} (\uV - \torque), \label{eq:torque_eom} \end{align}
which augments the inverted pendulum dynamics from \eqref{eq:actpend_eom_compact}. Here, $\omega_\text{b} = 2 \pi f_\text{b}$ is the identified actuation bandwidth.

The swing-up trajectory is obtained by solving a discrete-time optimal control problem over a finite horizon $N = T/T_\text{s}$, with the actuation model included and a user-defined fixed transition time $T = \unit[0.6]{s}$:
\begin{subequations}
\begin{align}
    \min_{\uV} \quad & \sum_{k=0}^{N-1} \|\uV[k]\|_2^2 \\
    \text{s.t.} \quad & \xtauV[k+1] = \mathbf{f}_d(\xtauV[k], \uV[k]), \label{eq:discrete_dynamics} \\
    & \xtauV[0] = [\alpha_0, \varphi_0, 0, 0, \tau_{0}]^\intercal, \\
    & \xtauV[N] = \mathbf{0}, \quad \uV[N-1] = 0, \\
    & \ddot{\actangle}[N] = 0, \quad \ddot{\pendangle}[N] = 0, \\
    & |\uV[k]| < \bar{\uV}, \quad \forall k \in \{0, \dots, N-1\}.
\end{align}
\end{subequations}
Here, the augmented state vector $\xtauV = [\actangle, \pendangle, \dot{\actangle}, \dot{\pendangle}, \tau]^\intercal$ evolves according to the nonlinear discrete time dynamics $\mathbf{f}_d(\cdot)$, obtained using 4th-order Runge-Kutta discretization of the combined mechanical \eqref{eq:actpend_eom_compact} and torque \eqref{eq:torque_eom} dynamics. The initial condition includes a gravity compensation torque $\tau_{0}$, derived from \eqref{eq:actpend_eom_compact} under static conditions, i.e., $\dot{\actangle}=\dot{\pendangle}=\ddot{\actangle}=\ddot{\pendangle}=0$, which yields:
\begin{equation}
    \tau_{0} = -\Pact \gravity \sin{\alpha_0} -\Ppend \gravity \sin{\pendangle_0}.
\end{equation}
This term simulates the support of the physical holder, preventing the optimizer from assuming a free-fall condition at the trajectory onset.

The optimization problem is solved using the CasADi \cite{andersson2019} library with the iterative IPOPT solver \cite{wachter05}. 
% The solver is initialized with:
% \begin{subequations}
% \begin{align}
%     \xtauV[k] &= [(1 - k/N)\actangle_0, (1 - k/N)\pendangle_0, 0, 0, -\bar{\uV}]^\intercal, \\
%     \uV[k] &= -\bar{\uV},
% \end{align}
% \end{subequations}
% where the states are defined for $k \in \{0, \dots, N\}$ and the inputs for $k \in \{0, \dots, N-1\}$. 
To ensure accurate integration of the soft constraint dynamics, 4th-order Runge-Kutta integration is employed. 

\begin{figure}
    \centering
    \includegraphics[width=\columnwidth]{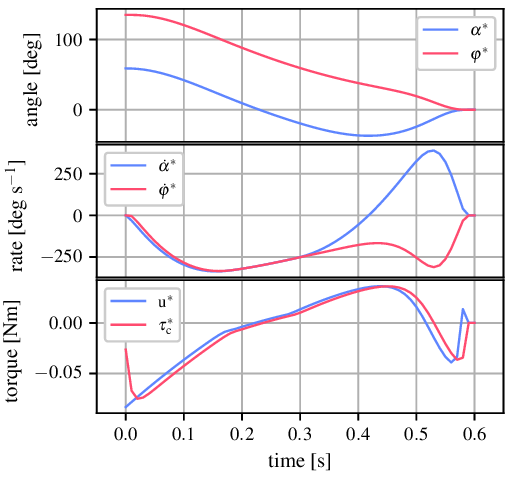}
    \caption{Time-series evolution of the optimal swing-up trajectory. Top \& Middle: The actuator ($\dot{\actangle}$) and pendulum ($\dot{\pendangle}$) angular rates remain coincident for $t < \unit[0.3]{s}$, confirming that the system initially accelerates as a pseudo-rigid body with the unilateral hard-stop engaged. The subsequent divergence at $t \approx \unit[0.3]{s}$ marks the inertial detachment phase where the pendulum is driven toward the upright equilibrium. Bottom: The simulated actual magnetic torque $\torque^*$ tracks the commanded setpoint $\utrajV$ with a visible first-order lag, consistent with the modeled actuation bandwidth. The trajectory respects the hard-stop constraint through the soft approximation, with a maximum penetration of $2.7^\circ$.}
    \label{fig:reference_trajectory}
\end{figure}

The resulting reference trajectory, shown in \cref{fig:reference_trajectory}, resembles previously reported swing-up trajectories in \cite{turrisi20, turrisi22, eom15}, with the key distinction that it explicitly exploits the mechanical hard stop to execute the maneuver. 

\subsection{Trajectory Tracking Feedback}
\begin{figure}
    \centering
    \includegraphics[width=\columnwidth]{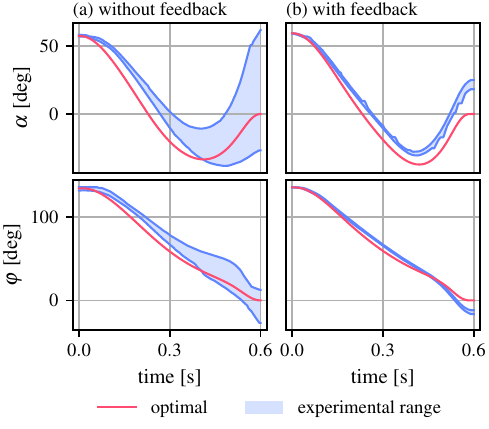}
    \caption{Experimental comparison of trajectory tracking consistency between open-loop and closed-loop control strategies. (a) Open-loop response performed on the OctoMag system (17 trials). The system exhibits significant sensitivity to initial conditions and disturbances, resulting in a final state dispersion (half-range) of approximately $\pm 44.1^\circ$ for $\actangle$ and $\pm 19.8^\circ$ for $\pendangle$. (b) Closed-loop response performed on the Navion system (11 trials). The feedback controller demonstrates high repeatability, reducing the final state dispersion to $\pm 3.4^\circ$ for $\actangle$ and $\pm 2.2^\circ$ for $\pendangle$. Note that although different electromagnetic navigation systems (eMNS) and reference trajectories were used for (a) and (b), the results are directly comparable as the proposed modeling framework decouples the magnetic field generation from the rigid-body dynamics, making the control strategy hardware-agnostic.}
    \label{fig:w_wo_fb_comparison}
\end{figure}

Under open-loop control, where the system evolves solely under the reference input trajectory $\utrajV$ without state-dependent compensation, experimental observations revealed a significant sensitivity to initial conditions and unmodeled disturbances, as illustrated in \cref{fig:w_wo_fb_comparison}a. To reject these non-repetitive disturbances, we employ a finite-horizon discrete-time LQR controller with the feedback control law:
\begin{align}
    \uV_{\text{FB}}[k] = \KtrajM[k] \tildexV[k],
\end{align}
where $\KtrajM[k]$ is the time-varying gain matrix calculated by solving the discrete-time Riccati equation associated with \eqref{eq:discretized} recursively starting at the timestep $N$. As demonstrated in \cref{fig:w_wo_fb_comparison}b, the feedback controller substantially reduces trajectory dispersion, enabling consistent performance across trials. However, it alone is insufficient to achieve the swing-up maneuver, as the system fails to reach the upright equilibrium close enough to trigger the switch to the balancing controller.

\subsection{Iterative Learning Controller}

To overcome the remaining tracking error, we deploy an ILC scheme that learns how to swing up the inverted pendulum. The tracking error arises from systematic modeling mismatches, including imperfect magnetic field calibration, unmodeled joint friction, and hard stop approximation errors. Crucially, these mismatches manifest consistently across trials, as illustrated in \cref{fig:w_wo_fb_comparison}b.

We exploit this repeatability using ILC, a powerful iterative technique that systematically improves performance by refining feedforward commands based on data from previous executions and the analytical model of the system \cite{bristow06}. Beyond modeling errors, the repetitive structure of physiological disturbances such as heartbeat and respiration makes ILC also promising for eMNS-guided clinical interventions \cite{cagneau07, wang24, bebek07}.

In the following, we detail the ILC scheme employed in this work, which is applied to the closed-loop system stabilized by the feedback controller, as shown in \cref{fig:control_loop}. Our methodology combines optimal filtering with convex optimization and is closely inspired by \cite{schoellig12}. For the $j$-th trial, we stack the variables into a lifted representation over the finite horizon $N$:
\begin{subequations}
\begin{align}
    % \superxV &= [ \tildexV[1], \dots, \tildexV[N] ]^\intercal \in \mathbb{R}^{Nn_x} \\
    % \superyV &= [ \tildeyV[1], \dots, \tildeyV[N] ]^\intercal \in \mathbb{R}^{Nn_y} \\
    \superuV &= [ \uV_{\text{ILC}}[0], \dots, \uV_{\text{ILC}}[N-1] ]^\intercal \in \mathbb{R}^{Nn_u}
\end{align}
\end{subequations}
where $\uV_{\text{ILC}}[k]$ represents the learned feedforward correction signal, which is added to the closed-loop torque command, as shown in \cref{fig:control_loop}.

The relation between the lifted state and input in the iteration domain is given by:
\begin{align}
    \superxV_j = \liftoutM (\liftinM \superuV_j + \distV_j ) + \boldsymbol{\mu}_j,
\end{align}
where the term $\boldsymbol{\mu}_j$ captures non-repetitive measurement noise, assumed to be zero-mean Gaussian white noise with covariance $\boldsymbol{M}$: $\boldsymbol{\mu}_j \sim \mathcal{N}(\zeroM, \boldsymbol{M})$. The lifted input matrix $\liftinM \in \mathbb{R}^{N n_x \times N n_u}$ is composed of sub-matrices $\liftinM_{(l,m)} \in \mathbb{R}^{n_x \times n_u}$, $1 \leq l, m \leq N$:
\begin{align}
    \liftinM = \begin{bmatrix} 
    \liftinM_{(1,1)} & \cdots & \liftinM_{(1,N)} \\
    \vdots & \ddots & \vdots \\
    \liftinM_{(N,1)} & \cdots & \liftinM_{(N,N)}
    \end{bmatrix},
\end{align}
with:
\begin{align}
    \liftinM_{(l,m)} = \begin{cases} 
    \clstateM[l-1] \cdots \clstateM[m] \inputM[m-1] & \text{if } m < l, \\
    \inputM[m-1] & \text{if } m = l, \\
    0 & \text{if } m > l,
    \end{cases}
\end{align}
where $\clstateM[k]$ is the time-varying closed-loop state transition matrix, defined as:
\begin{align} 
    \clstateM[k] = \stateM[k] + \inputM[k]\KtrajM[k].
\end{align}
The repetitive disturbance $\distV_j$ is assumed to evolve as a random walk across trials:
\begin{align}
    \distV_j = \distV_{j-1} + \boldsymbol{\omega}_{j-1},
\end{align}
where the process noise $\boldsymbol{\omega}_{j-1} \sim \mathcal{N}(\zeroM, \boldsymbol{\Omega})$ accounts for trial-to-trial variations of the disturbance.

Fundamentally, the goal is to identify the repetitive disturbance $\distV_j$ and compensate for it in the subsequent iteration through the feedforward correction input $\superuV_{j+1}$. To achieve this, the disturbance is estimated using an iteration-domain Kalman filter:
\begin{align}
     \distestV_j = \distestV_{j-1} + \kalmanM_j (\superyV_j - \liftoutM \distestV_{j-1} - \liftoutM \liftinM \superuV_j),
\end{align}
where $\kalmanM_j$ is the Kalman gain. The optimal feedforward correction for the subsequent trial is then found by solving the quadratic program:
\begin{equation}
\begin{aligned}
    \min_{\superuV_{j+1}} \quad & \left\| \scaleM \left( \liftinM \superuV_{j+1} + \distestV_j \right) \right\|_2^2 + \xi \left\| \fdM \superuV_{j+1} \right\|_2^2 \\
    \text{s.t.} \quad & |\uV[k]| \leq \bar{\uV}, \quad \forall k \in \{0, \dots, N-1\},
\end{aligned}
\label{eq:ILC_input_update}
\end{equation}
where $\scaleM \in \mathbb{R}^{N n_x \times N n_x}$ is a diagonal weighting matrix, $\xi$ is a regularization parameter \cite{schoellig12}, and $\fdM \in \mathbb{R}^{N n_u \times N n_u}$ is a finite-differences matrix defined as:
\begin{align}
    \fdM = \frac{1}{T_\text{s}}\begin{bmatrix} 
    -1 & 1 & & \\
    & \ddots & \ddots & \\
    & & -1 & 1
    \end{bmatrix}.
\end{align}
The first term in the cost function penalizes the residual trajectory error after applying the feedforward correction $\superuV_{j+1}$, while the second term allows us to penalize rapid changes in the correction signal to ensure smoothness.

Since the linearization from \eqref{eq:discretized} holds only when the actual trajectory remains close to the reference trajectory, significant deviations invalidate the lifted matrices $\liftinM$ and $\liftoutM$, compromising both the disturbance estimation and input update steps. We thus employ a horizon extension strategy \cite{schoellig12}. During each trial, we monitor the state deviation from the reference trajectory. If this deviation exceeds a predefined threshold at time step $k$, the trial is terminated early at index $N_j \leq N$. The ILC optimization is then solved only over the valid segment $k \in [0, N_j]$, producing an updated input for this portion. In subsequent trials, improved tracking typically allows the system to progress further before termination, gradually extending the learned horizon as performance improves across iterations.

\begin{figure}[htbp]
    \centering
    \includegraphics[width=\columnwidth]{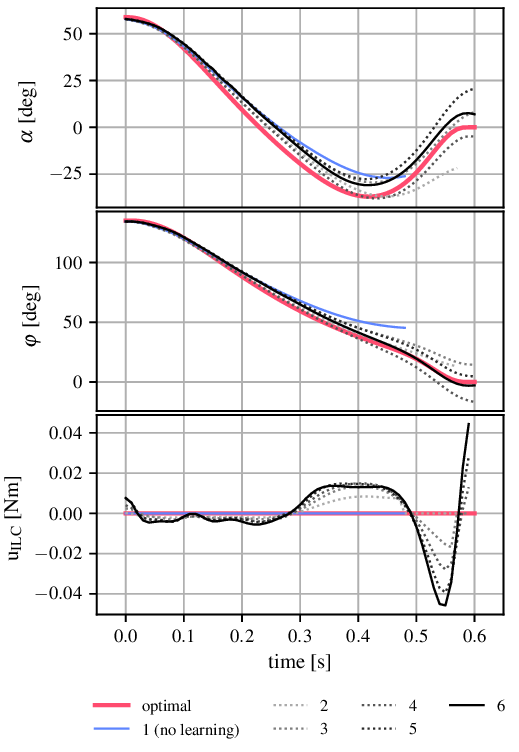}
    \caption{ILC learning progression for the swing-up maneuver over six iterations on the Navion eMNS. Actuator angle $\actangle$ (top), pendulum angle $\pendangle$ (middle), and feedforward correction signal $\uV_{\text{ILC}}$ (bottom) show progressive refinement from initial failure (iteration 1) to successful trajectory tracking (iteration 6) relative to the reference trajectory (red).}
    \label{fig:iterations}
\end{figure}

As shown in \cref{fig:iterations}, over five subsequent input updates, the ILC scheme progressively refined the feedforward input and extended the valid execution horizon. By the sixth iteration, the system closely tracked the reference trajectory into the switching region, upon which the balancing controller engages to balance the pendulum in the upright equilibrium.

\subsection{Balancing Controller}
Once the state enters the empirically defined region of attraction $\mathcal{S}$, defined by the thresholds:
\begin{subequations}
\begin{align}
    |\actangle|      &\leq 30^\circ,       \quad &|\dot{\actangle}| &\leq 300^\circ/\text{s}, \\
    |\pendangle|     &\leq 10^\circ,       \quad &|\dot{\pendangle}| &\leq 100^\circ/\text{s}.
\end{align}
\end{subequations}
the trajectory tracking phase terminates and the balancing controller engages to balance the pendulum in the upright equilibrium. To this end, we employ an infinite-horizon discrete-time LQR regulator with the control law:
\begin{align}
    \uV_{\text{eq}}[k] = \KeqM \xV[k],
\end{align}
where $\KeqM$ is the constant gain matrix obtained by solving the discrete-time algebraic Riccati equation associated with \eqref{eq:discretized} at the upright equilibrium $\xtrajV = [0,0,0,0]^\intercal$ \cite{zughaibi25a}.

\section{Experimental Results}
\label{sec:results}

\cref{fig:iterations} illustrates the progression of the ILC scheme over six iterations. Under feedback-only control, the system fails to track the reference trajectory due to systematic modeling errors, resulting in early termination. Over subsequent iterations, the learned feedforward correction progressively compensates for these mismatches, with the control input converging as the learning progresses. By the sixth iteration, the system successfully tracks the reference trajectory throughout the swing-up maneuver, entering the predefined region of attraction and triggering the switch to the balancing controller.

\begin{figure*}[!ht]
    \centering
    \includegraphics[width=2\columnwidth]{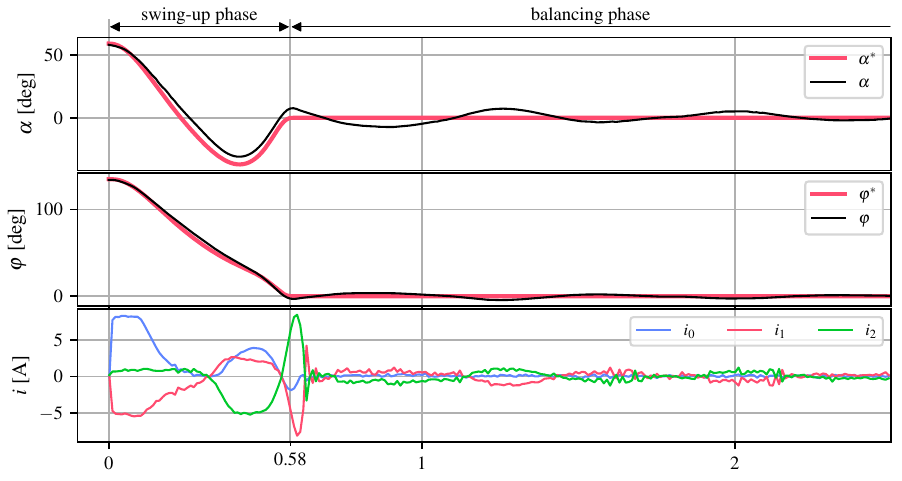}
    \caption{Trajectory tracking and stabilization results for the final ILC iteration on the Navion eMNS. The actuator angle $\actangle$ (top) and pendulum angle $\pendangle$ (center) closely track their respective reference trajectories $\actangle^*$ and $\pendangle^*$ (red) during the swing-up phase. After controller switching at $t = \unit[0.58]{s}$, the system successfully stabilizes the pendulum in the upright configuration. The bottom panel displays the coil currents of Navion, which remain strictly within the \unit[13]{A} safety threshold throughout the maneuver.}
    \label{fig:final_iteration}
\end{figure*}

Upon switching to the balancing controller, the system successfully stabilized the pendulum in the upright configuration, as shown in \cref{fig:final_iteration}. We observed slight residual limit-cycle oscillations near equilibrium, which we attribute to joint friction \cite{campbell08, zughaibi18}. Despite these oscillations, the successful swing-up demonstrates that ILC effectively compensated for modeling mismatches.

\begin{figure}
    \centering
    \includegraphics[width=\columnwidth]{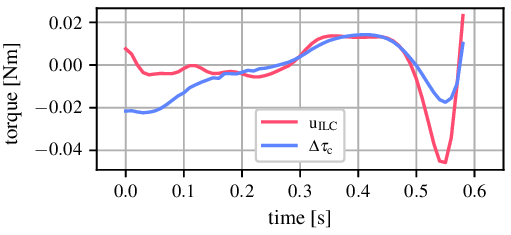}
    \caption{Comparison of the final ILC input correction $\uV_{\text{ILC}}$ and the calculated torque discrepancy $\Delta\torque$, defined as the difference between the torque calculated using the initial calibration and a high-fidelity post-experimental calibration of the magnetic field model for the Navion eMNS. Both signals exhibit a correlation starting from $t = \unit[0.15]{s}$ and are truncated at $t = \unit[0.58]{s}$, where the controller switches from the trajectory tracking to the balancing controller.}
    \label{fig:torque_discrepancy}
\end{figure}

In particular, the post-experimental analysis revealed that the learned feedforward correction accounted for significant discrepancies in the initial magnetic field model. To quantify this mismatch, we computed the torque discrepancy $\Delta\torque$ by evaluating \eqref{eq:torqueforce_projection} with the measured coil currents $\currentsV$, actuator angle $\actangle$, and dipole position $\positionV$ for both the initial and high-fidelity calibration obtained from \cite{bernardes26}:
\begin{align}
\Delta\torque = \jacM(\actangle)\dipoleM(\actangle)\left(\actuationM(\positionV) - \actuationM_{\text{hf}}(\positionV)\right)\currentsV,
\end{align}
where $\actuationM$ and $\actuationM_{\text{hf}}$ denote the actuation matrices obtained from the initial and high-fidelity post-experimental calibrations, respectively. As shown in \cref{fig:torque_discrepancy}, the ILC correction signal closely matches this predicted torque discrepancy with Pearson's correlation coefficient $r=0.933$ starting at $t=\unit[0.15]{s}$, suggesting that learning-based approaches can effectively handle the inherent field model uncertainties in the electromagnetic actuation. Note that the approach is general and extends beyond field model uncertainties to other repetitive disturbances inherent to electromagnetic navigation, such as physiological motions induced by heartbeat and respiration in cardiac ablation procedures \cite{cagneau07, wang24, bebek07}.

% These findings suggest that learning-based approaches can effectively handle inherent uncertainties in electromagnetic actuation, with implications for clinical applications of eMNS. Moreover, the repetitive error structure that ILC exploits is characteristic of physiological disturbances such as heartbeat and respiration, which could be leveraged in procedures such as cardiac ablation, where recurring perturbations induced by patient motion are difficult to model explicitly \cite{cagneau07, wang24, bebek07}.

% This framework could further extend to multi-agent magnetic manipulation, where complex inter-agent coupling effects are difficult to model analytically but manifest consistently across repeated task executions, making them natural candidates for iterative compensation.

\section{Conclusion}
\label{sec:conclusion}

In this work, we presented a platform-agnostic framework for non-equilibrium trajectory tracking of underactuated magnetic systems, demonstrating the first swing-up of a magnetically actuated inverted pendulum on a clinically-ready eMNS. Successfully executing this maneuver showcases the high actuation bandwidth of eMNS, proving that highly dynamic control is feasible even at a clinical scale. By combining bandwidth-aware trajectory optimization, time-varying LQR feedback, and ILC, we showed that systematic model mismatches arising from imperfect field calibration and unmodeled mechanical effects can be effectively compensated through iterative feedforward refinement, achieving successful swing-up within six ILC iterations.

In addition, post-experimental analysis revealed that the learned ILC correction closely matches the torque discrepancy predicted by a subsequent high-fidelity field calibration, suggesting learning and adaptation as a promising tool to handle uncertainties inherent to electromagnetic actuation. More broadly, the repetitive error structure exploited by ILC is also characteristic of physiological disturbances such as heartbeat and respiration, suggesting a natural pathway toward robust eMNS control during clinical interventions such as cardiac ablation.

Future work includes evaluating the proposed control framework against real physiological disturbances. Furthermore, the approach can be extended to multi-agent scenarios, where model uncertainties arise from magnetic interactions between the permanent magnets of different agents and cross-coupling through the shared electromagnetic coils \cite{zughaibi25b}. %Applying this framework to multi-agent magnetic manipulation may enable surgical knot tying, an inherently repetitive multi-tool task involving coordinated motion of multiple agents.

\section*{Acknowledgements}

The authors thank António Bernardes for his assistance with magnetic field calibration and Thomas Steinbrenner for his hardware design support. Michael Muehlebach acknowledges financial support from the German Research Foundation, and Jasan Zughaibi thanks the Max Planck ETH Center for Learning Systems and the Swiss National Science Foundation (Grant IZLCZ0\_206033). Additionally, the authors utilized generative AI assistance for sentence restructuring to improve the manuscript's readability.

\section*{Conflict of Interest}

The authors declare no conflicts of interest.

%% Use plainnat to work nicely with natbib. 

% \bibliographystyle{plainnat}
\bibliographystyle{unsrtnat}
\bibliography{references}

\appendices

\section{Implementation Details}

This section provides implementation details complementing the control framework described in \cref{sec:control}. All control and learning parameters are summarized in \cref{tab:hyperparameters}.

\begin{table}[ht]
\centering
\caption{LQR and ILC Hyperparameters}
\label{tab:hyperparameters}
\begin{tabular}{@{}lll@{}}
\toprule
 & Parameter & Value \\
\midrule
\multirow{3}{*}{Trajectory tracking LQR} 
    & $\mathbf{Q}_\text{t}$ & diag(1, 1, 0.01, 0.01) \\
    & $R_\text{t}$ & $10^5$ \\
    & $\mathbf{Q}_\text{t,f}$ & diag(1, 1, 0.01, 0.01) \\
\midrule
\multirow{2}{*}{Balancing LQR \cite{zughaibi25a}} 
    & $\mathbf{Q}_\text{eq}$ & diag(1, 1, 0.1, 0.1) \\
    & $R_\text{eq}$ & $10^4$ \\
\midrule
\multirow{5}{*}{ILC} 
    & $\mathbf{P}_0$ & $\mathbf{I}$ \\
    & $\boldsymbol{M}$ & $0.01 \cdot \mathbf{I}$ \\
    & $\boldsymbol{\Omega}$ & $0.05 \cdot \mathbf{I}$ \\
    & $\xi$ & 0.05 \\
    & $\distV_0$ & $\zeroM$ \\
\bottomrule
\end{tabular}
\end{table}

To ensure validity of the linearized dynamics used in the ILC law, a trial is aborted if either the rod deviates from the reference trajectory by more than $\unit[20]{^\circ}$, as the first-order approximation of the nonlinear dynamics degrades significantly beyond this threshold \cite{schoellig12}. The scaling matrix $\scaleM$ normalizes all states to the range $[-1, 1]$ along the reference trajectory, thereby accounting for the different orders of magnitude between angular positions and velocities in \eqref{eq:ILC_input_update}. Finally, to address the system's inherent latencies, an actuation delay of approximately \unit[40]{ms} is compensated by pre-applying the initial feedforward torque $\uV_{\text{ff}}(0)$ and holding it constant for the duration of the delay period prior to trajectory execution.

Following \cite{zughaibi25a}, a constant feedforward torque offset is applied during balancing to compensate for steady-state errors arising from motion capture calibration misalignments and actuation matrix inaccuracies. This offset is determined prior to the ILC iterations by running the balancing controller with integral action until convergence and storing the resulting integral torques as a fixed feedforward term.

\end{document}